\documentclass[unnumsec,webpdf,modern,large,namedate]{oup-authoring-template}

\usepackage{booktabs}
\usepackage{array}

\usepackage{caption}

\theoremstyle{thmstyleone}%
\theoremstyle{thmstyletwo}%
\theoremstyle{thmstylethree}%

\renewcommand{\cite}[1]{\citep{#1}}

\begin{document}

\journaltitle{Accepted to OUP Bioinformatics Advances}
\DOI{https://doi.org/10.1093/bioadv/vbae117}
\copyrightyear{2024}
\appnotes{} 

\firstpage{1}


\title[ENBED]{Understanding the Natural Language of DNA using Encoder-Decoder Foundation Models with Byte-level Precision}

\author[1, 2, $\ast$]{Aditya Malusare}
\author[2]{Harish Kothandaraman}
\author[3]{Dipesh Tamboli}
\author[2, 4]{Nadia A. Lanman}
\author[1, 2, 3]{Vaneet Aggarwal}

\authormark{Malusare et al.}

\address[1]{\orgdiv{School of Industrial Engineering}, \orgname{Purdue University, USA}}
\address[2]{\orgdiv{Institute for Cancer Research, Purdue University, USA} }
\address[3]{\orgdiv{Elmore Family School of Electrical and Computer Engineering}, \orgname{Purdue University, USA}}
\address[4]{\orgdiv{Department of Comparative Pathobiology}, \orgname{Purdue University, USA}}

\corresp[$\ast$]{Corresponding author. \href{email:malusare@purdue.edu}{malusare@purdue.edu}}




\abstract{This paper presents the Ensemble Nucleotide Byte-level Encoder-Decoder (ENBED) foundation model, analyzing DNA sequences at byte-level precision with an encoder-decoder Transformer architecture. ENBED uses a sub-quadratic implementation of attention to develop an efficient model capable of sequence-to-sequence transformations, generalizing previous genomic models with encoder-only or decoder-only architectures. We use Masked Language Modeling to pre-train the foundation model using reference genome sequences and apply it in the following downstream tasks: (1)  identification of enhancers, promotors and splice sites, (2) recognition of sequences containing base call mismatches and insertion/deletion errors, an advantage over tokenization schemes involving multiple base pairs, which lose the ability to analyze with byte-level precision, (3) identification of biological function annotations of genomic sequences, and (4) generating mutations of the Influenza virus using the encoder-decoder architecture and validating them against real-world observations. In each of these tasks, we demonstrate significant improvement as compared to the existing state-of-the-art results. }


\maketitle
%
%
%

\newcommand{\modelname}[0]{ENBED\space}

\section{Introduction}

The rise of foundation models in recent years has led to tremendous developments in understanding natural languages \cite{Paa2023}. Although they were originally developed to process and generate written text, these models have transcended their initial purpose due to their generalizable nature and wide applicability. Foundation models have shown great potential in the field of bioinformatics \cite{Zhang2023}, since their capacity to be trained on vast amounts of unlabeled data and their adaptability enable them to achieve state-of-the-art performance in a variety of tasks.

Early applications of foundation models in bioinformatics can be seen in analyzing protein sequences \cite{intro1, intro2}, which were then trained on diverse applications like calculation of protein structure, prediction of mutation effects and the understanding of phylogenetic structure \cite{Lupo2022ProteinLM, Fang2022AMF, Nijkamp2022ProGen2ET}. These models have since evolved beyond proteins into DNA and RNA analysis, and have demonstrated the ability to surpass previous benchmarks in identifying regulatory elements, predicting chromatin profiles, analyzing evolution from genomic sequence data and predicting the impacts of mutations in DNA \cite{jiDNABERTPretrainedBidirectional2021, DallaTorre2023TheNT, Nguyen2023HyenaDNALG, Zvyagin2022GenSLMsGL, Yamada2021PredictionOR}. The ability to visualize and interpret the internal model structure \cite{Vig2020BERTologyMB} and to derive key insights of the underlying biological processes \cite{zhangTransformerGeneExpression2022}  demonstrate the unique advantages offered by foundation models in the field of bioinformatics. 






\subsection{Limitations of previous work}

\subsubsection{Architecture. }

Prior work on Transformer-based models for DNA sequence analysis exists in two forms: (i) Encoder-only models \cite{jiDNABERTPretrainedBidirectional2021,fishmanGENALMFamilyOpenSource2023, zhangTransformerGeneExpression2022, DallaTorre2023TheNT} that focus on classification and regression-based downstream tasks and (ii) Decoder-only models \cite{Nguyen2023HyenaDNALG, zhangDNAGPTGeneralizedPretrained2023} that are capable of classification, regression as well as generative tasks that involve design and synthesis. 

 A combination of encoder and decoder blocks enables the model to perform sequence-to-sequence transformations. One of the fundamental processes undergone by DNA is its transcription into an RNA sequence and subsequent translation into protein sequences, the building blocks of all living organisms. Understanding sequence-to-sequence processes like these is crucial to advancing our knowledge of genetics, and developing an encoder-decoder model is an important step in this direction. Although decoder-only models are capable of sequence-to-sequence transformations, they have no independent means of creating representations of the input sequence, and both input and target tokens are processed in an equivalent fashion. {\color{black} Previous work has shown that a multitask finetuned encoder-decoder Large Language Model (LLM) outperforms decoder-only models on zero-shot generalization \cite{sanh2022multitask} as well as targeted tasks like machine translation \cite{raffelExploringLimitsTransfer2020,fu2023decoderonly}}. Since a decoder-only architecture will have a unidirectional framework that attends to the source and target sequence simultaneously, as the length of the target sequence grows, the extent to which the model attends to the source will decrease leading to reduced performance in downstream tasks \cite{fu2023decoderonly}. Our work demonstrates how the cross-attention layers in the decoder leverage the information in the embeddings generated by the encoder, leading to improved performance in training tasks. 

\subsubsection{Tokenization. }
Biological sequences like DNA are encoded using a vocabulary of four symbols (A, C, T, G) representing nucleic acids. These sequences are converted into a Transformer-compatible format by a tokenizer, which generates a list of tokens for any given input. Since these models were initially developed for applications in natural languages, the most prevalent forms of tokenization are sentence-piece or word-piece, where the language vocabulary is built using natural ideas like words or syllables. In the absence of typical indicators of linguistic order in DNA, like spaces and punctuation, these tokenization schemes use statistical techniques to determine the `words' that make up the vocabulary of the input sequences. A few examples of previously used tokenizers are: k-mer \cite{jiDNABERTPretrainedBidirectional2021}, SentencePiece \cite{DallaTorre2023TheNT}, and byte-pair encoding (BPE) \cite{fishmanGENALMFamilyOpenSource2023} tokenization. While such techniques identify optimal encoding methods by constructing tokens having multiple base pairs, they are vulnerable to any type of noise present in the sequence. A single variation in a base pair will result in the fragment being mapped to a completely different word in the vocabulary, resulting in an outsized impact from a small perturbation \cite{Dotan2023EffectOT}. We use a simplified tokenization scheme where each character corresponds to a single token, resulting in a longer average tokenized length, but more resiliance to the variations mentioned above. 

\subsection{Our contributions}

In this paper, we develop the Ensemble Nucleotide Byte-level Encoder-Decoder (\textbf{ENBED}) Transformer, a foundation model that analyzes nucleotide sequences with Transformers using byte-level tokenization and an encoder-decoder model. This implementation bridges the gap between existing models that are either encoder-only or decoder-only implementations and presents the possibility of sequence-to-sequence analysis tasks. Using sliding-window and global attention we obtain a sub-quadratic implementation of attention, and demonstrate the performance improvements over dense attention. The foundation model is pre-trained using an ensemble of high-quality reference genomes from NCBI RefSeq, including the telomere-to-telomere assemblies of Human and Maize DNA, data from the 1000 Genomes Project and a mix of widely studied organisms like \textit{E. coli}, \textit{D. melanogaster}, \textit{M. musculus} and \textit{P. vivax} (Sec \ref{sec:dataav}). This process is implemented by giving the model a self-supervised goal of internalizing the structure of the language of nucleotide sequences. 


\modelname is built using a byte-level tokenizer. In order to avoid the issues created by single nucleotide variants and their downstream impacts, we side-step the problem of determining the tokenization scheme entirely by working with single nucleotides as tokens. This leads to increased computational costs, but grants resilience to the types of variations and noise commonly encountered in DNA sequences. In order to offset the impact of increased computations, we implement sub-quadratic attention layers in order to scale up the model efficiently. 

\subsubsection{Evaluation of performance on genomic benchmark datasets.} We evaluate the performance of the \modelname foundation model on sequence-level classification tasks and compare it's accuracy against contemporary foundation models. We show that \modelname outperforms the state-of-the-art in 21 of the 25 benchmarks devised by the authors of the Nucleotide Transformer \cite{DallaTorre2023TheNT} and Genomic Benchmarks \cite{Greov2022GenomicBA} datasets. These benchmarks consist of tasks like identifying enhancers, promotors, splice sites and histone marks in multi-species data comprising of genomic sequences from human, mouse, yeast, fruit fly and worm DNA. 

\subsubsection{Identifying sequencing noise.} Long-read sequencing using Nanopores is used to study telomeres, which are protective caps found at chromosomal ends and have long repetitive elements. It has been found that telomeres in many organisms are frequently miscalled \cite{Tan2022}, referring to errors in the process that translates electrical signals into the alphabet of DNA. We illustrate how \modelname can focus on fragments that look incorrect or out of place, demonstrating the model's ability of distinguishing between noisy and accurate data. In a synthetic dataset constructed using noise distributions found in real-world raw sequence data, we demonstrate that our model can identify sequences containing noise with an accuracy of 97.6\%, leveraging the information internalized by bring pretrained on the telomere-to-telomere reference sequences.  

\subsubsection{Biological function annotations.} Mapping the complete human genome was a significant milestone in modern biology, and it has produced a new set of challenges in identifying the functions and interactions of different parts of the genome. We fine-tune our model to solve a version of this problem by identifying the biological functions of genomic sequences among the most common functional classes using a fine-tuned model, achieving an $F_1$ score of 74.1.

\subsubsection{Studying mutations as a sequence-to-sequence process.} 
Exploring mutations is essential as it sheds light on the mechanisms driving genetic diversity which enhance the overall resilience of living organisms in a changing environment. The encoder-decoder architecture confers the ability to rapidly iterate mutagenization of genomic segments. We study mutations in the Influenza virus, using the NCBI Influenza Virus Resource. By constructing a dataset with a phylogenetic tree, we obtain parent-child pairs of mutated sequences and show the effectiveness of our encoder-decoder architecture in analyzing and predicting these mutations. 

\section{Methods}

\subsection{Encoder-Decoder Model Architecture}
\label{sec:arch}

\begin{figure*}[!t]
\centering
	\includegraphics[width=\textwidth]{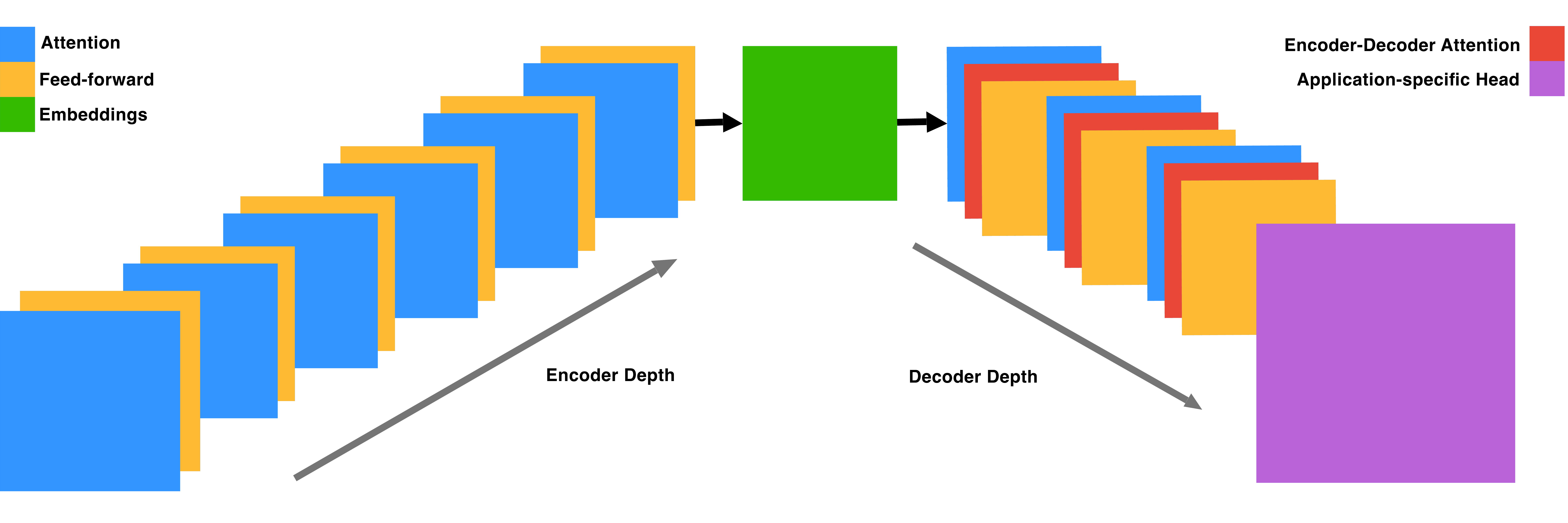}
	\caption{\textbf{Model Architecture.} The model is constructed using encoder and decoder blocks with a  ratio of 2:1. Both types of blocks consist of attention and feed-forward layers, with the decoder blocks additionally incorporating the embeddings in encoder-decoder attention layers.  }
	\label{fig:modelarch}	
\end{figure*}

\modelname is built using an encoder-decoder architecture (Fig. \ref{fig:modelarch}) consisting of encoder and decoder blocks, each comprised of two subcomponents: an attention layer and a feed-forward neural network. {\color{black} The attention layers process a sequence by replacing each element with a weighted sum of linear transformations of the input embeddings}, after which they are normalized and passed through the feed-forward neural network. Dropout is applied to the feed-forward network, the attention weights, and the input and output of the entire stack. The implementation is written using JAX \cite{jax2018github} and the Flax-former library \cite{flax2020github}. 

We formulate a model with 1.2B trainable parameters, with the configuration specified in Table \ref{table:modelparams}. The model is encoder-heavy since idiosyncratic relationships among tokens are better encoded by devoting a larger share of parameters to these blocks. We found that adjusting the encoder-to-decoder ratio to 2:1 improved performance, with a ~1\% increase in Masked Language Modeling (MLM) accuracy for all model sizes over the 3:1 ratio chosen by the authors of ByT5 \cite{xueByT5TokenfreeFuture2022}, a similar architecture built to process token-free text-to-text transformations. We also find that reducing the masked span length, which is the average number of tokens masked during pre-training, from 40 down to 20 helps in faster convergence owing to the significantly smaller vocabulary of DNA.

\subsection{Tokenization}

Sequences are tokenized by breaking down the input into tokens consisting of single nucleotides. The vocabulary size is fixed at 384, with 256 ASCII characters and additional tokens added to function as \verb|MASK|, \verb|PAD| and \verb|UNKNOWN| tokens during the training process. We require multiple \verb|MASK| tokens in order to index the positions where masking has occurred and to label the targets with these indices. Although the alphabet of DNA only comprises of the four nucleic acids Adenine (A), Cytosine (C), Guanine (G), and Thymine (T), we choose to keep the whole set of extended ASCII characters since they could aid in future tasks like sequence-to-sequence transformations involving targets beyond just DNA sequences, like drug structures represented by the SMILES notation system. 

This approach requires more floating-point operations (FLOPs) as compared to other tokenization methods, since it increases the tokenized sequence length for the same input DNA sequences, resulting in higher resource requirements. Although this limits us to dealing with short- to medium-length sequences, we can overcome these constraints and scale up the model by  reducing the complexity of attention layers as described below. 

\subsection{Attention} 
\label{sec:attention}

Attention  can be understood as a soft-lookup of a query $\textbf{Q}$ in a dictionary of stored keys $\textbf{K}$ and values $\textbf{V}$. Attention scores are generated by calculating the similarity between $\textbf{Q}$ and $\textbf{K}$, each having a dimension $d$, with scaled dot-product attention $\left(\text{Softmax}\left(\textbf{Q}\textbf{K}^T/\sqrt{d}\right)V\right)$ being the most common implementation. Increasing the sequence length $L$ can be a challenge, since this type of attention has a complexity of $O(L^2)$. This sets a limit of $L\leq 512$ tokens on our hardware (NVIDIA A100 (40 GB) GPUs). 

In order to reduce the complexity while preserving function, we modify the architecture to replace dense attention with a combination of two sub-quadratic variants of attention: (i) sliding-window attention and (ii) global attention. 

\subsubsection{Sliding-window attention. } 
Local context is crucial in analyzing DNA, since biological processes like transcription and translation work within continuous regions of a sequence. Tokens within a sliding window of radius $r$ are used to calculate the attention scores, bringing the complexity down to $O(L\times r)$. We fix $r=64$ for the initial three layers and increase to $r=128$ in the final layers, which allows them to learn higher-level representations while having the lower layers focus on local information. 

\subsubsection{Global attention. }
For tasks that involve classifying or annotating whole sequences, we need a mechanism that aggregates global information from the inputs, in addition to the local scores. We divide the input sequence into $k$ blocks and calculate a global token by summing and normalizing the embeddings for every token in the block. Scores are then computed for every input token by letting it attend to the neighboring tokens (as described above) and all the global tokens, which has a total complexity of $O(L(r + k))$. 
\\
\\
Hence, by choosing appropriate values for $r$ and $k$ relative to $L$, we implement a scheme to calculate attention with a sub-quadratic complexity which allows us to set an input and output length of $16384$, a significant improvement over the limit of $512$ tokens using dense attention with the same GPU hardware. 

{\color{black} The aggregated blocks constructed in this procedure resemble previous tokenization schemes like k-mer, used by previous models like DNABERT \cite{jiDNABERTPretrainedBidirectional2021} and BPE used by GENA-LM \cite{fishmanGENALMFamilyOpenSource2023}. Our method uses a combination of these aggregated blocks along with higher-granularity local context to achieve a balance between the two, allowing us to process sequences with greater precision.}

\subsection{Applications of Foundation Models using Transfer Learning}

\subsubsection{Building the foundation model.}
\label{sec:pretrain}
The first step in building our foundation model is pre-training it on high-quality reference sequences. We use a procedure called Masked Language Modeling (MLM). The objective is to reconstruct tokens that have been deleted and replaced with a \verb|MASK| token. This task develops the ability to understand the context and vocabulary to identify the correct elements that belong in the masked segments. Utilizing a large corpus of unlabeled data allows us to impart the model with generalizable knowledge that can be fine-tuned for specific downstream tasks. The genomic corpus is constructed by concatenating FASTA files from the NCBI sources mentioned in the Data Availability section, removing any descriptions starting with `$>$' and `N' bases that are a result of hard-masking. We choose a masking rate of $15\%$ over the course of pre-training. The entire corpus is supplied to a collator that handles masking, padding, and truncation to ensure that the input length is maintained. We follow a linear schedule with warmup (5\% of the total training steps) using the AdamW optimizer ($\beta_1=0.9, \beta_2=0.99, \epsilon=10^{-6}$) with a learning rate of 1e-5, a cross-entropy loss function and softmax as the activation function. We train all versions of the model with maximum input and output lengths of $16,384$ tokens (base pairs). Convergence takes 120-480 GPU-hours with 8 NVIDIA A100 GPUs, determined by model size and input length.


\subsubsection{Fine-tuning for downstream tasks.}
\label{sec:finetune}

We fine-tune the model by modifying the final layers into a task-specific configuration. This is called the `head' of the model and is attached to the final layer of the pre-trained model. Layers are gradually unfrozen in reverse order during the course of fine-tuning, allowing the Transformer to integrate with the attached head while retaining the initial layers, thus enabling the transfer of pre-trained knowledge for downstream applications. 

\subsubsection{Classification head. }
A fully connected (dense) layer is usually added to the output of the base model, followed by a softmax activation to produce class probabilities, typically used in sequence-level classification tasks. 

\subsubsection{Language modeling head. } A language modeling head comprises of a single feedforward neural network layer followed by a softmax activation function. This layer takes hidden representations from the preceding layers and outputs a probability distribution over the vocabulary. The objective is to estimate the estimate the probability of a token given the previous words in a sentence. The softmax function transforms the raw output scores into probabilities, representing the likelihood of each word or token in the vocabulary at any particular position. This process is called autoregressive generation, and we use it to perform sequence-to-sequence transformations.

\subsection{Application Domains}
\label{sec:applicationdomains}

{\color{black} The ENBED foundation model is evaluated across a set of genomic analysis tasks to demonstrate its versatility and the unique advantages of its encoder-decoder architecture. We begin with the Genomic Benchmarks and Nucleotide Transformer Benchmarks, which provide standardized comparisons against existing models for fundamental sequence classification tasks. The noise identification task assesses ENBED's ability to distinguish genuine sequences from artifacts, leveraging its byte-level precision. Biological function annotation tests the model's capacity to associate sequence patterns with higher-level functions, crucial for genome interpretation. Finally, the mutation generation task is an end-to-end evaluation of the ENBED, a novel architecture not present in previous genomic language models. This sequence-to-sequence task, focused on predicting viral mutations, showcases ENBED's potential for modeling complex genomic transformations.}

\subsubsection{Genomic Benchmarks. }  The Genomic Benchmarks (GB) dataset consists of sequences from four organisms: Human, mouse (\textit{Mus musculus}), roundworm (\textit{Caenorhabditis elegans}) and fruit fly (\textit{Drosophila melanogaster}). The dataset comprises of: (i) Human enhancers from Cohn et. al. \cite{Cohn2018EnhancerIU} and Ensembl \cite{Martin2022Ensembl2}, (ii) Open Chromatin Region classifications from the Ensembl build, (iii) Computationally generated data for coding and non-coding sequences (iv) Multi-class data composed of three regulatory elements (promotors, enhancers and Open Chromatin Regions), (v) Non-TATA promotor sequences imported from Umarov et. al. \cite{umarov2017recognition}.

\subsubsection{Nucleotide Transformer Benchmarks.} The Nucleotide Transformer (NT) benchmarks consist of five data sources: (i) Epigenetic marks in the yeast genome, which use experimentally obtained nucleosome occupancy values processed into positive and negative observations and to provide the following histone marks datasets: \{H3, H4, H3K9ac, H3K14ac, H4ac, H3K4me1, H3K4me2, H3K4me3, H3K36me3, and H3K79me3\}, (ii) A dataset  \cite{geng2022deep} consisting of a mix of strong, weak and non-enhancers. (iii) Promotor sequences $300$ base pairs in length around transcription start sites, divided on the basis of TATA and non-TATA box promotors. (iv) Splice site datasets composed of donor, acceptor and non-splice site sequences from phylogenetically diverse organisms. 

\subsubsection{Noise identification.}

We generate a synthetic dataset with segments of $512$ nucleotides selected at random from TeloBase \cite{Lycka2023}, a comprehensive database of information about telomere motif diversity. We introduce noise based on real-world raw DNA sequencing data to generate negative samples. {Previous work \color{black} \cite{Rabadan2017} finds that noise in sufficiently deep DNA sequencing data can be approximated by aggregating negative binomial distributions. Using this method, we create a balanced dataset with positive and negative samples.} The model is fine-tuned on a sequence classification task with this labeled dataset. This process can be likened to out-of-distribution detection \cite{Fort2021ExploringTL}, since the negative samples would represent data that does not belong to the distribution of the training dataset. We describe this procedure in more detail in the Supplementary Material (Section B).

\subsubsection{Biological function annotation.} 
We can formulate the process of annotating genes as a classification task, with the input being a DNA sequence fragment and the output {\color{black} being the class probabilities for the annotation types defined below}. For evaluating our model, we train it to output the biological function annotation of a given genomic input sequence up to 512 base pairs in length. {\color{black} We choose the following annotation types for our experiment: Coding Sequences, IncRNA, snoRNA, miscRNA, miRNA, snRNA, TEC, Processed and Unprocessed Pseudogenes. These annotations are obtained from the Ensembl dataset \cite{Martin2022Ensembl2}, and the constructed dataset has an equal number of examples for all classes. We generate 9216 training examples and 1024 validation examples for this task.}

\subsubsection{Mutation generation.} 

Human influenza A viruses are named based on the geographic location where the virus was isolated, the date of the isolate, and the identity of the two major surface proteins, hemagglutinin (HA) and neuraminidase (NA). We choose the HA1 sequences to create the Influenza virus mutation dataset, selecting the segments with most highly variable regions for training and validation. We obtain our source data from  \cite{Berman2020MutaGANAS} and subset the HA1 nucleotide sequence of the H3N2 Influenza virus between 300 to 799 bp (100-266 amino-acids) to capture the Antigenic site A and B. The selected region is a part of the globular domain that occurs in a jelly-roll fold of eight-stranded anti-parallel beta-sheets, containing the most commonly mutating amino-acid residues around the receptor binding site. The HA1 head also accumulates N-linked glycosylation sites over time, which are thought to mask antigenic sites from immune recognition. The glycosylation of the HA1 globular domain modulates receptor binding, stimulates host antibody responses, and shields key antigenic sites to facilitate immune evasion of the virus. By focusing on the HA1 subdomain, we aimed to evaluate the sequence-to-sequence model on a functionally important region of influenza HA that experiences significant antigenic drift and glycosylation changes. The Supplementary Material contains additional details about the construction of training and validation splits for the dataset.

Candidate sequences are generated using a language modeling head with the parent sequence supplied as the input. Using a beam search ($N_{beams}$=5), we obtain five candidate sequences which are autoregressively generated to a length of 499 bp (equal to the input). We rank the sequences using the noise identification pipeline above, and select the sequence least likely to be identified as having noise present. We identify mutations by measuring the Levenshtein distance between parent and child sequences. This metric accounts for insertion, deletion as well as in-place modifications.

\section{Results}

\begin{figure*}[!t]
    \centering
    \begin{center}
    \includegraphics[width=0.7\textwidth]{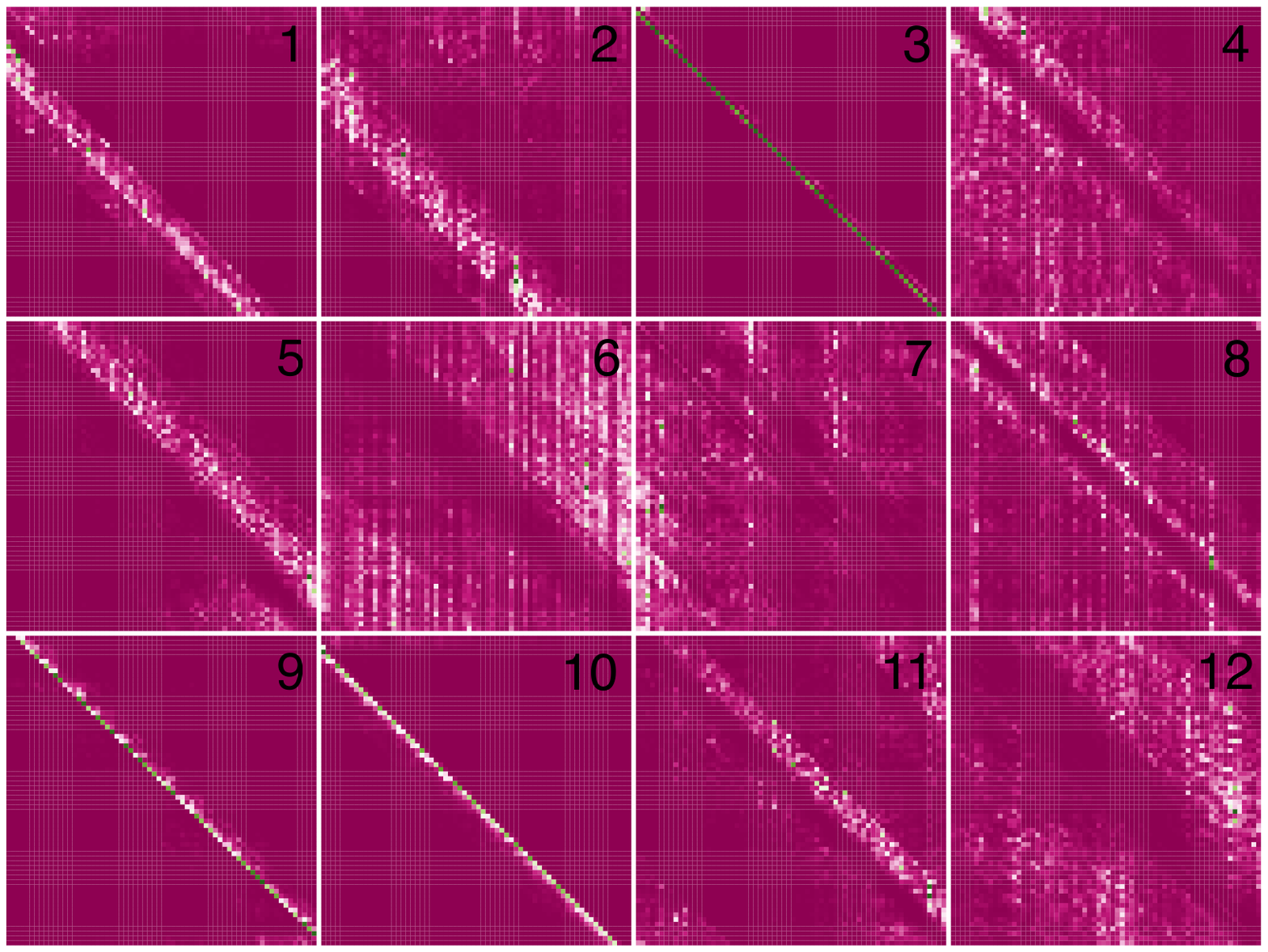}
    \includegraphics[width=0.5\textwidth]{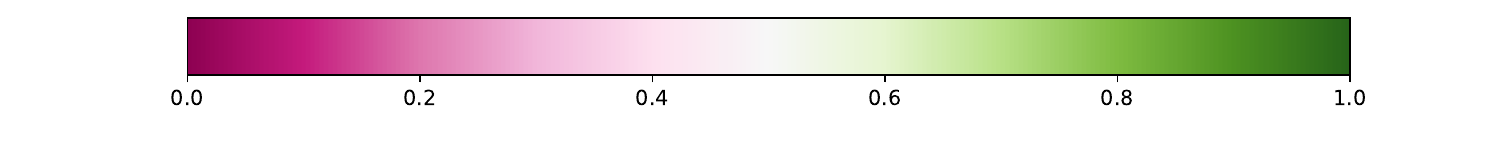} 
    \end{center}
    \caption{\textbf{Interpreting Attention Layers.} We visualize the twelve attention heads of the pre-trained \modelname foundation model.}
    \label{fig:maps}

\end{figure*}

Upon convergence, the pre-training process yields a foundation model ready to be applied to downstream tasks. The initial layers in the pre-trained model are frozen since they contain generalizable information that helps the model build versatile internal representations of the data. We visualize these internal representations by extracting the encoder output layer and plotting attention maps in Fig. \ref{fig:maps}. These maps are generated using the outputs from the final encoder block. The use of multiple attention heads grants the model the ability to simultaneously use a diverse range of patterns to analyze input sequences. In Fig \ref{fig:maps}, we observe that some heads are dedicated to analyzing close neighbors (3, 9, 10) while others display a more dilated version of this phenomenon (1, 2, 5, 11). Additionally, there are heads which attempt to exclude local information and focus on a more global view of the input sequence (4, 6, 8, 12).

\subsection{\modelname outperforms state-of-the-art models on genomic benchmark datasets}
\label{sec:benchmarks}

We finetune the model using a classification head using the embedding outputs from the final encoder block, on the datasets constructed by the authors of the Nucleotide Transformer (NT) benchmarks \cite{DallaTorre2023TheNT} and Genomic Benchmarks (GB) \cite{Greov2022GenomicBA}.  The results of evaluating the model on the test dataset of NT and GB are presented in Tables \ref{table:ntbenchmarks} and \ref{table:genbenchmarks}, respectively. For evaluation on the NT benchmarks, we compare our performance against the Nucleotide Transformer (v2) and HyenaDNA \cite{DallaTorre2023TheNT, Nguyen2023HyenaDNALG}, which are encoder-only and decoder-only models,  respectively. For the GB datasets, we use the performance of the Convolutional Neural Network (CNN) model developed by the authors of the dataset \cite{Greov2022GenomicBA} as a baseline. We also include the performance of the HyenaDNA model and the baseline Transformer developed by its authors \cite{Nguyen2023HyenaDNALG}.

{\color{black} ENBED demonstrates superior performance, exceeding state-of-the-art results in 15 out of 17 Nucleotide Transformer (NT) benchmarks and 6 out of 8 Genomic Benchmarks (GB) datasets. This improvement likely stems from our novel approach combining byte-level analysis, high-quality reference sequences, and an optimized pre-training methodology. We hypothesize that byte-level tokenization enhances the model's ability to handle variations such as single nucleotide polymorphisms, while our encoder-decoder architecture enables simultaneous focus on multiple input sections and context-aware processing. These features may contribute to ENBED's advantages over decoder-only methods. While the relative impact of each component requires further investigation through ablation studies, our results demonstrate ENBED's effectiveness across a wide range of genomic analysis tasks.}

\subsection{\modelname identifies noise in genomic sequences}

Table \ref{table:errorID} shows the results of the sequence-level classification on erroneous sequences using our synthetic dataset. Since competing models are trained using the GRCh38 reference assembly, they often lack information about repetitive regions due to hard-masking. Our choice of higher quality pre-training data results in a signifcant performance improvement and on overall accuracy of 97.1\% in the sequence-level classification task of identifying erroneous genomic data, which is significant improvement as compared to the baselines of DNABERT \cite{jiDNABERTPretrainedBidirectional2021}  (84.9\%) and Nucleotide Transformer \cite{DallaTorre2023TheNT} (91.8\%).

\subsection{\modelname identifies biological function annotations}
\label{sec:codingpot}

{\color{black}\modelname is trained to identify the annotations (defined in the Application Domains section) of the Human reference assembly.} As shown in Table \ref{table:biotype}, we achieve an $F_1$ score of 74.1 in this classification task, an improved score  compared to DNABERT \cite{jiDNABERTPretrainedBidirectional2021} (63.2), Nucleotide Transformer \cite{DallaTorre2023TheNT} (67.5), and HyenaDNA \cite{Nguyen2023HyenaDNALG} (72.8). {\color{black} For the purposes of this evaluation, all models were finetuned and evaluated using the same balanced dataset as specified in the Application Domains section.}

\subsection{\modelname generates mutations using sequence-to-sequence transformation}
\label{sec:geneannotation}

We evaluate the accuracy of \modelname in generating mutations, using an encoder-decoder Transformer with Byte-Pair Encoding (BPE) tokenization (used in previous genomic models \cite{fishmanGENALMFamilyOpenSource2023}) as a baseline. We compare against BPE because this method shares similarities with byte-level tokenization by starting with the basic \{A, C, T, G\} alphabet, but tries to optimize the vocabulary by combining simpler words into more complex ones based on the corpus the tokenizer is trained on. {\color{black} The training corpus itself is identical to the one used in pre-training ENBED, with the only difference being the tokenization procedure}. While this procedure reduces the average number of tokens generated from any input sequence, it also results in reduced accuracy since modifying even a single base pair will output a significantly different tokenized sequence.

Top-1 and Top-5 Accuracy (\%) scores are calculated by comparing predictions with real-world data from the Influenza Virus Resource \cite{Bao2008}, with any deviation from an exact match being classified as incorrect. Top-5 scores are calculated by selecting the best candidate from the procedure described in Sec \ref{sec:applicationdomains}. Additionally, we also train a version of ENBED with the encoder removed, as a comparison of the sequence-to-sequence task performance between decoder-only and encoder-decoder models. 

The mean Levenshtein distance of our model predictions from real-world mutated sequences is 2.3 edits over a length of 500 bp, resulting in an average similarity of 99.5\%. We can attribute the significant increase in accuracy to byte-level tokenization, since other schemes with tokens involving multiple base pairs will be unable to capture edits involving single nucleotides effectively.

\section{Ablation Studies}

We perform ablation studies in order to examine the impact of the architectural modifications and the combination of encoder and decoder blocks. 

\subsection{Encoder-decoder architecture}

We study the impact of combining encoder and decoder blocks and the cross-attention links between them in Table \ref{table:modelparams}. A decoder-only version of the model is constructed by stacking 24 decoder layers and is pre-trained to convergence using next-token prediction. We also construct a balanced model using stacks of 12 layers for both the encoder and decoder blocks, introducing cross-attention layers in the decoder that attend to the embeddings and the output sequence. Both models have $\sim$ 800 M trainable parameters. We then fine-tune these models on the mutation generation task and compare with the ENBED model having a 2:1 encoder-decoder block ratio. 

Introducing the encoder and cross attention leads to a significant improvement in the pre-training accuracy, demonstrating the suitability of both the architecture as well has the pre-training task, since decoder-only models are restricted to causal objectives like next-token prediction unlike encoders that can handle bi-directional information.  

{\color{black}
\section{Discussion}

The ENBED model demonstrates significant improvements over existing approaches in several areas of genomic sequence analysis. The encoder-decoder architecture, combined with byte-level tokenization and high-quality pre-training data, contributes to enhanced performance across multiple tasks. ENBED's performance on established benchmarks is noteworthy, surpassing state-of-the-art results in 21 out of 25 tasks across the Nucleotide Transformer and Genomic Benchmarks datasets. This broad improvement suggests that our approach captures underlying genomic patterns more effectively than previous models. Additionally, the model successfully identified sequences containing noise with an accuracy of 97.6\%, demonstrating its sensitivity to small-scale genomic perturbations. This is likely due to the byte-level tokenization approach used in ENBED, which allowed for accurate detection of variations at single-nucleotide resolutions.

The encoder-decoder structure proves particularly effective for sequence-to-sequence tasks like mutation generation. Our results show that ENBED outperforms baseline models in predicting Influenza virus mutations, achieving a top-5 accuracy of 95.4\%. This was a significant improvement over the baseline model using byte-pair encoding (BPE) tokenization (56.1\%), and another variant of ENBED without the encoder (72.1\%). We chose to vary both the tokenization scheme and architecture in these cases while keeping the rest of the design choices unchanged in order to isolate the impact of these two factors. We find that the choice of BPE tokenization significantly impacts the model's ability to generate mutations accurately, with byte-level tokenization providing a clear advantage due to its ability to capture single-nucleotide changes. We also see that an encoder-decoder architecture is crucial for this task, as the decoder-only model does not perform as well, following the trend observed in other sequence-to-sequence tasks \cite{raffelExploringLimitsTransfer2020,fu2023decoderonly}. 

It is also worth noting that the use of higher-quality pre-training data, including telomere-to-telomere assemblies, may contribute to ENBED's improved performance. This comprehensive genomic representation likely allows the model to learn from previously underrepresented genomic regions. A study of the Nucleotide Transformer benchmarks (Table \ref{table:ntbenchmarks}) with two versions of ENBED trained on different reference assemblies (GRCh38 and T2T-CHM13) showed that the model trained on the higher-quality T2T-CHM13 assembly outperformed the GRCh38 model across the board. This suggests that the choice of reference assembly can significantly impact the model's performance, and that the use of more complete and accurate reference genomes can lead to better generalization.

Future work on this model could explore additional applications in genomics, such as variant effect prediction and protein structure studies.  

}
\section{Data and Code Availability}
\label{sec:dataav}

The Telomere-to-telomere reference sequences for Human 
(GCF\_009914755.1)
 and Maize 
 (GCA\_022117705.1) and the reference sequences for \textit{E. coli} (GCF\_000008865.2), \textit{D. melanogaster} (GCF\_000001215.4), \textit{M. musculus} \newline(GCF\_000001635.27) and \textit{P. vivax} (GCF\_000002415.2) were obtained from NCBI RefSeq \cite{OLeary2015} in FASTA format. Variant Calling Files (VCFs) for the 1000 Genomes Project \cite{tgp} were obtained from the European Bioinformatics Institute. Gene annotations were obtained from GENCODE \cite{GENCODE} and Ensembl \cite{Martin2022Ensembl2}. The mutation tree was derived from the data assembled by the authors of \cite{Berman2020MutaGANAS}, sourced from the NCBI’s Influenza Virus Resource \cite{Bao2008}. 
\\
The source code used to develop and fine-tune the foundation model has been released on Github \footnote{https://github.itap.purdue.edu/Clan-labs/ENBED} and the weights of the model used in evaluation are available here \footnote{https://huggingface.co/malusare}

\section{Supplementary Material}

The supplementary material (below) contains additional details on the model architecture, data sources, training procedures and evaluation metrics.

\section{Acknowledgements}
This work was supported in part by the National Science Foundation under grant [FW-HTF-R-2129097]; and the Anvil supercomputer \cite{anvil} at Purdue University through allocation CIS230228 from the Advanced Cyberinfrastructure Coordination Ecosystem: Services \& Support (ACCESS) program, which is supported by National Science Foundation grants [2138259, 2138286, 2138307, 2137603, and 2138296].  The authors gratefully acknowledge the Walther Cancer Foundation and support from the Purdue University Institute for Cancer Research, [P30CA023168].

\begin{table*}
\centering
\begin{center} 
\begin{tabular}{c c c c c | >{\color{black}}c >{\color{black}}c}

\toprule

\textbf{NT Benchmark} & Enformer  & DNABERT-2 & \begin{tabular}{@{}c@{}}NT \\ (2.5B)\end{tabular}   &  \begin{tabular}{@{}c@{}}HyenaDNA \\ (1 Kb)\end{tabular}   & \begin{tabular}{@{}c@{}}ENBED \\ (GRCh38)\end{tabular}  & ENBED  \\
\midrule
H3                    & 0.719     &  0.785     & \underline{0.791}	& 0.779         & 0.723 &\textbf{0.802} \\
H3K14ac               & 0.288     &  0.516     & 0.537  & \underline{0.612}	 		& 0.537 & \textbf{0.636} \\
H3K36me3              & 0.344     &  0.591     & \underline{0.616}	& 0.613         & 0.611 & \textbf{0.624} \\
H3K4me1               & 0.291     &  0.511     & \underline{0.544}	& 0.512         & 0.498 & \textbf{0.591} \\
H3K4me2               & 0.211     &  0.336     & 0.322  & \underline{0.455} 		& 0.433 & \textbf{0.501} \\
H3K4me3               & 0.158     &  0.352     & 0.408  & 0.549 		& \underline{0.580} & \textbf{0.587} \\
H3K79me3              & 0.496     &  0.613     & 0.621  & \underline{0.672} 		& 0.648 & \textbf{0.756} \\
H3K9ac                & 0.420     &  0.542     & 0.550  & \underline{0.581} 		& 0.427 & \textbf{0.590} \\
H4                    & 0.732     &  0.796     & \underline{0.807}	& 0.763         & 0.750 & \textbf{0.823} \\
H4ac                  & 0.273     &  0.463     & 0.489  & \underline{0.564} 		& 0.548 & \textbf{0.605} \\
\midrule
Promotor (all)        & 0.909     & 0.943      & \underline{0.950}  & 0.920         & 0.906 & \textbf{0.961} \\
Promotor (non-TATA)   & 0.909     & 0.944      & \underline{0.952}  & 0.921         & 0.892 & \textbf{0.959} \\
Promotor (TATA)       & 0.920     & 0.910      & \underline{0.919}  & 0.882         & 0.883 & \textbf{0.944} \\
\midrule
Splice acceptor       & 0.829     & \underline{0.950}      & \textbf{0.973}  & 0.915			& 0.754 & 0.943  \\
Splice donor          & 0.814     & \underline{0.926}      & \textbf{0.974}  & 0.898			& 0.835 & 0.911  \\
\midrule
Enhancer              & 0.451     & 0.516      & 0.548  & 0.517         & \underline{0.577} & \textbf{0.585} \\
Enhancer Types        & 0.309     & 0.423      & 0.450  & 0.386         & \underline{0.459} & \textbf{0.482} \\
\bottomrule
\vspace{2mm}
\end{tabular}

\caption{\textbf{Nucleotide Transformer (NT) Benchmarks.} We evaluate our model using the {\color{black} 10-fold mean Matthews Correlation Coefficient (MCC)} of the best performing variants of the Enformer \cite{Avsec2021}, DNABERT \cite{zhou2023dnabert2}, Nucleotide Transformer v2 \cite{DallaTorre2023TheNT}, and HyenaDNA \cite{Nguyen2023HyenaDNALG}, highlighting the \textbf{best} and \underline{second-best} scores. The scores are sourced from a leaderboard maintained by the authors of \cite{DallaTorre2023TheNT} on the Hugging Face platform \cite{leaderboard}.}

\label{table:ntbenchmarks}
\end{center}
\end{table*}

\begin{table*}[ht]
\centering
\setlength{\tabcolsep}{4pt}
\centering
\begin{center} 
\begin{tabular}{c c c c c | >{\color{black}}c c}

	\hline
	\noalign{\vskip 1mm}
	\textbf{Genomic Benchmark}&CNN & DNABERT & GPT  &\begin{tabular}{@{}c@{}}HyenaDNA \\ \cite{Nguyen2023HyenaDNALG}\end{tabular} & \begin{tabular}{@{}c@{}}ENBED \\ (GRCh38)\end{tabular} & \modelname \\
	\noalign{\vskip 1mm}
	\hline
	\noalign{\vskip 1mm}

Mouse Enhancers            & 69.0     & 66.9   & 80.1         & \underline{85.1}  & 81.1 &  \textbf{90.3}             \\
Human Enhancers (Cohn)     & 69.5     & \underline{74.0}   & 70.5         & \textbf{74.2}  & 70.8 & {71.2}            \\
Human Enhancers (Ensembl)  & 68.9     & 85.7   & 83.5         & \underline{89.2}  & 90.2 & \textbf{92.2}              \\
Coding vs Intergenomic     & 87.6     & \underline{92.5}   & 88.8 & 91.3          & 90.7 & \textbf{93.0}              \\
Human vs Worm              & 93.0     & {96.5}   & 95.6         & \underline{96.6}  & 94.4 & \textbf{97.3}            \\
Human Regulatory Elements          & \underline{93.3} & 88.1 & 91.5         & \textbf{93.8} & 85.6 &  90.2            \\
Human Promoter (Non-TATA)   & 84.6    & 85.6     & 87.7         & \underline{96.6}  & 90.4 & \textbf{97.2}            \\
Human OCR (Ensembl) & 68.0       &    75.1   & 73.0         & \underline{80.9}  & 76.2 & \textbf{81.9}                \\
	\noalign{\vskip 1mm}
	\hline
	\noalign{\vskip 2mm}
\end{tabular}

\caption{\textbf{Genomic Benchmarks.} Accuracy (\%) scores of the \textbf{best} and \underline{second-best} model in the Genomic Benchmarks datasets \cite{Greov2022GenomicBA}. The baseline CNN and GPT scores was calculated by the authors of \cite{Greov2022GenomicBA} and \cite{Nguyen2023HyenaDNALG} respectively.}

\label{table:genbenchmarks}
\end{center}
\end{table*}

{\color{black}

\begin{table}
\setlength{\tabcolsep}{4pt}
\centering
\begin{center} 
\begin{tabular}{c c c c}

	\hline
	\noalign{\vskip 1mm}
	\textbf{Model}&Reference&$F_1$ Score\\
	\noalign{\vskip 1mm}
	\hline
	\noalign{\vskip 1mm}
	DNABERT&\cite{jiDNABERTPretrainedBidirectional2021}&84.9\\
	Nucleotide Transformer&\cite{DallaTorre2023TheNT}&91.8\\
	\noalign{\vskip 1mm}
	\hline
	\noalign{\vskip 1mm}
	\textbf{\modelname}& This paper &{\color{black} \textbf{97.6}}\\
	\noalign{\vskip 1mm}
	\hline
	\noalign{\vskip 2mm}

\end{tabular}

\caption{\textbf{Erroneous Sequence Identification.} }
\label{table:errorID}
\vspace{-2mm}
\end{center}
\end{table}

}

\begin{table}

\setlength{\tabcolsep}{4pt}
\centering
\begin{center} 
\begin{tabular}{c c c c}

	\hline
	\noalign{\vskip 1mm}
	\textbf{Model}&Reference&$F_1$ Score\\
	\noalign{\vskip 1mm}
	\hline
	\noalign{\vskip 1mm}
	DNABERT&\cite{jiDNABERTPretrainedBidirectional2021}&63.2\\
	Nucleotide Transformer&\cite{DallaTorre2023TheNT}&67.5\\
	HyenaDNA&\cite{Nguyen2023HyenaDNALG}&72.8\\
	\noalign{\vskip 1mm}
	\hline
	\noalign{\vskip 1mm}
	\textbf{\modelname}& This paper &\textbf{74.1}\\
	\noalign{\vskip 1mm}
	\hline
	\noalign{\vskip 2mm}

\end{tabular}

\caption{\textbf{Biological Function Identification.} }
\label{table:biotype}
\end{center}
\end{table}

\begin{table*}
\setlength{\tabcolsep}{4pt}
\centering
\begin{center} 
\begin{tabular}{c c c c c}

	\hline
	\noalign{\vskip 1mm}
	\textbf{Model}&Top-1 Accuracy&Top-5 Accuracy&Mean LD&Median LD\\
	\noalign{\vskip 1mm}
	\hline
	\noalign{\vskip 1mm}
	Transformer (BPE tokenization)&32.0&56.1&30.6&24\\
    ENBED (decoder-only)&53.1&72.1&6.1&4\\
    \modelname &\textbf{76.9}&\textbf{95.4}&2.3&1\\
	\noalign{\vskip 1mm}
	\hline
	\noalign{\vskip 2mm}

\end{tabular}

\caption{\textbf{Mutation Generation.} Accuracy (\%) scores of Top-1 and Top-5 candidates with the mean and median Levenshtein Distance (LD) between predicted and child sequences.}
\label{table:mutation}
\end{center}

\end{table*}

\begin{table}
\setlength{\tabcolsep}{3pt}
\centering
\begin{center} 
\begin{tabular}{c c c c}

	\hline
	\noalign{\vskip 1mm}
	\textbf{Configuration}&\begin{tabular}{@{}c@{}}Decoder-only \\ (no Cross-Attn.)\end{tabular} & 
                \begin{tabular}{@{}c@{}}Base model \\ 1:1 Enc/Dec\end{tabular}&ENBED\\
	\noalign{\vskip 1mm}
	\hline
	\noalign{\vskip 1mm}
	Parameters&800M&800M&1.2B\\
	$d_{ff}$&3584&3584&3850\\
	$d_{kv}$&64&64&64\\
	$d_{model}$&1536&1536&1536\\
	Encoder layers&0&12&24\\
	Decoder layers&24&12&12\\
	Attention heads&16&16&16\\
	 Global attention ($k$)&128&128&256\\
	\noalign{\vskip 1mm}
	\hline
	\noalign{\vskip 1mm}
	Top-1 accuracy (\%)&53.1&62.0&76.9\\
	\noalign{\vskip 1mm}
	\hline
	\noalign{\vskip 2mm}
\end{tabular}
\caption{\textbf{Model Configurations}. $d_{model}$ denotes the size of the encoder layers, and the pooler layer, $d_{kv}$ is the size of the key, query, and value projections per attention head and $d_{ff}$ is the size of the intermediate feed-forward layer in each Transformer block. The accuracy of the top-1 candidate is evaluated using the same framework used in Table \ref{table:mutation}.}
\label{table:modelparams} 
\end{center}
\end{table}


\newpage

\bibliography{manual,zotero}
\bibliographystyle{plainnat}

\onecolumn

\clearpage
\section{Supplementary Material}	

\begin{appendices}
	
\renewcommand{\cite}[1]{\citep{#1}}

\section{ Pre-training Data Sources }

Table \ref{table:pretraining} shows the pre-training data sources used for the Enformer \citep{Avsec2021}, DNABERT-2 \citep{zhou2023dnabert2}, Nucleotide Transformer v2 \cite{DallaTorre2023TheNT}, and HyenaDNA \cite{Nguyen2023HyenaDNALG} models. We also construct a GRCh38-based version of ENBED as mentioned in Tables 1 and 2 in the main paper. 

\begin{table*}[b]
\centering
\begin{center}
\begin{tabular}{ccc}
\toprule
\textbf{Model} & \textbf{Data Source} & \textbf{Description} \\
\midrule
Enformer \cite{Avsec2021} & GRCh38 + GRCm38 & Human and Mouse reference genomes  \\
DNABERT-2 \cite{zhou2023dnabert2} & \begin{tabular}{@{}c@{}} GRCh38 + \\ Multi-species Dataset \end{tabular}& 
\begin{tabular}{@{}c@{}} Multi-species data consists of 135 species \\ randomly selected across 7 categories. \end{tabular} \\
\begin{tabular}{@{}c@{}} Nucleotide Transformer \\ \cite{DallaTorre2023TheNT} \end{tabular}  & \begin{tabular}{@{}c@{}} GRCh38 + 1000G + \\ Multi-species Dataset \end{tabular} & \begin{tabular}{@{}c@{}} 
	Versions with the Human reference genome, \\ 1000 Genomes project (1000G), \\ and multi-species data consists of 850 species. \end{tabular}  \\
HyenaDNA \cite{Nguyen2023HyenaDNALG} & GRCh38 & Human reference genome \\
\bottomrule
\end{tabular}
\caption{\textbf{Pre-training Data Sources.} }
\label{table:pretraining}
\end{center}
\end{table*}

\section{ Task-specific Datasets }

\subsection{Nucleotide Transformer}

For epigenetic marks prediction, a dataset of acetylation and methylation nucleosome occupancies in the yeast genome was used, with data from Chip-Chip experiments processed into positive and negative observations for 10 histone marks. Promoter sequence prediction utilized a dataset of 29,597 promoter regions, including 3,065 TATA-box promoters, with sequences spanning 300bp around transcription start sites. Matched negative samples were created by shuffling promoter sub-sequences.

Enhancer sequence prediction relied on a single dataset that originally contained 742 strong, 742 weak, and 1484 non-enhancers, which was augmented with 6000 synthetic enhancers and 6000 synthetic non-enhancers to evaluate the transformer's representation of enhancers. Splice site prediction employed two datasets: the SpliceFinder dataset, which included donor, acceptor, and non-splice sites in human genes with 400bp sequences, and the Spliceator training set, which consisted of 600bp sequences from diverse organisms, using a balanced 'Gold Standard' subset.

Table \ref{tab:dataset_stats}, sourced from Dalla-Torre et al. [2], shows the dataset statistics for the various genomic sequence classification tasks.

\begin{table}
	\centering
	\begin{tabular}{cccc}
\toprule
	 & Num train & Num test & Max sequence \\
	 & sequences & sequences & length in bp \\
\midrule
	 H3K4me3 & 25953 & 2884 & 500 \\
	H3K4me2 & 27614 & 3069 & 500 \\
	H3K36me3 & 31392 & 3488 & 500 \\
	H3K9ac & 25003 & 2779 & 500 \\
	Splice donor & 19775 & 2198 & 600 \\
	Splice site all & 27000 & 3000 & 400 \\
	H4ac & 30685 & 3410 & 500 \\
	H3K4me1 & 28509 & 3168 & 500 \\
	Enhancer & 14968 & 400 & 200 \\
	Enhancer types & 14968 & 400 & 200 \\
	H4 & 13140 & 1461 & 500 \\
	Splice acceptor & 19961 & 2218 & 600 \\
	H3K79me3 & 25953 & 2884 & 500 \\
	Promoter non-TATA & 47767 & 5299 & 300 \\
	Promoter all & 53276 & 5920 & 300 \\
	H3K14ac & 29743 & 3305 & 500 \\
	H3 & 13468 & 1497 & 500 \\
	Promoter TATA & 5509 & 621 & 300 \\
\bottomrule
\end{tabular}
	\caption{Dataset statistics for Nucleotide Transformer classification tasks}
	\label{tab:dataset_stats}
\end{table}

\subsection{Genomic Benchmarks}

The Genomic Benchmarks dataset consists of 8 classification tasks, each with a unique set of positive and negative sequences. The tasks include the classification of mouse enhancers, human enhancers (Cohn), human enhancers (Ensembl), coding vs. intergenic regions, human vs. worm, human regulatory elements, human promoters (non-TATA), and human OCR (Ensembl). The dataset is designed to evaluate the performance of models on a diverse set of genomic sequence classification tasks. Table \ref{tab:genomic_benchmarks} shows the dataset statistics for the Genomic Benchmarks tasks.

\begin{table}
	\centering
	\begin{tabular}{cccccc}
		\toprule
	Name & \# of sequences & \# of classes & Class ratio & Median length & $\sigma$ \\
		\midrule
	dummy\_mouse\_enhancers\_ensembl & 1210 & 2 & 1.0 & 2381 & 984.4 \\
	demo\_coding\_vs\_intergenomic\_seqs & 100000 & 2 & 1.0 & 200 & 0.0 \\
	demo\_human\_or\_worm & 100000 & 2 & 1.0 & 200 & 0.0 \\
	drosophila\_enhancers\_stark & 6914 & 2 & 1.0 & 2142 & 285.5 \\
	human\_enhancers\_cohn & 27791 & 2 & 1.0 & 500 & 0.0 \\
	human\_enhancers\_ensembl & 154842 & 2 & 1.0 & 269 & 122.6 \\
	human\_ensembl\_regulatory & 289061 & 3 & 1.2 & 401 & 184.3 \\
	human\_nontata\_promoters & 36131 & 2 & 1.2 & 251 & 0.0 \\
	human\_ocr\_ensembl & 174756 & 2 & 1.0 & 315 & 108.1 \\
		\bottomrule
\end{tabular}
	\caption{Description of datasets in genomic benchmark package. Name is the unique identification of dataset. \# of sequences is the combined count of all sequences from all classes. \# of classes is the count of all classes in a dataset. Class ratio is the ratio between number of sequences in the largest and smallest classes. Median length and Standard deviation are computed for all sequences from all classes in a dataset. (Reproduced from \cite{Greov2022GenomicBA})}
	\label{tab:genomic_benchmarks}
	\end{table}

\subsection{Noise Generation}

We generated a synthetic dataset to evaluate our model's capacity to differentiate between genuine sequences and those containing errors. The dataset was constructed using segments of 512 nucleotides selected at random from TeloBase, a comprehensive database of telomere motif diversity. 

Noise was injected as per the distribution found in the work of \cite{Rabadan2017} using a  deepSNV-based implementation \cite{gerstung2012reliable}. The dataset was divided into training and test sets with 10,000 and 1,000 sequences, respectively. 

\subsection{Mutation Generation}

For the mutation generation task, we employ a fine-tuning approach using a sequence-to-sequence model. This model is trained to predict child sequences given parent sequences, effectively learning the patterns of mutations observed in the influenza virus population. To ensure the robustness of our results and prevent overfitting, we have implemented a comprehensive strategy for constructing our training and test datasets.

\begin{figure}
    \centering
    \includegraphics[width=0.5\textwidth]{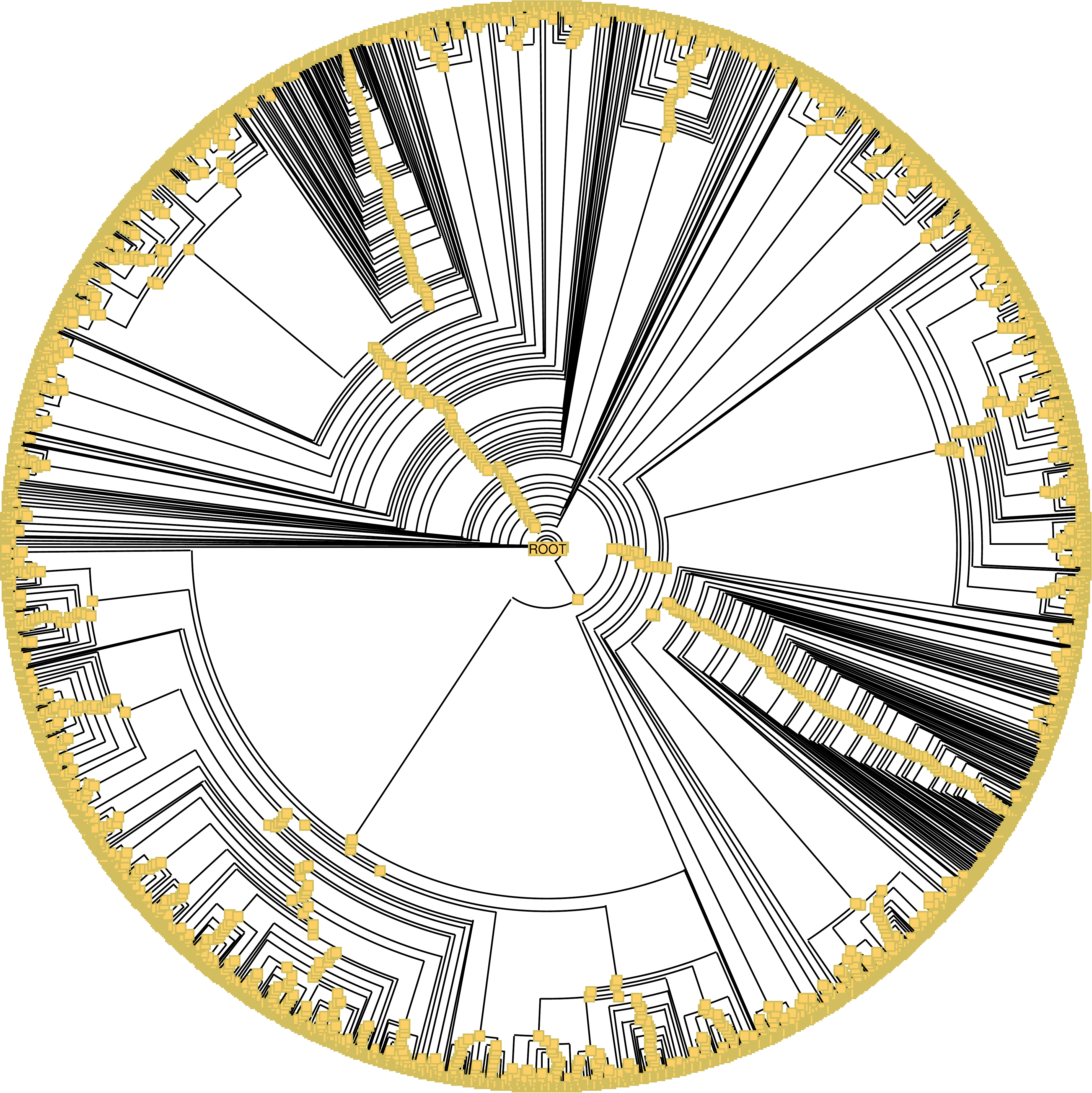}
    \caption{\textbf{Phylogenetic Tree.} }
    \label{fig:phylo}
\end{figure}

Our approach begins with the construction of a phylogenetic tree from the available influenza virus sequences using a maximum likelihood method. Figure \ref{fig:phylo} shows a circular cladogram visualization of the generated Influzenza H1 gene sequences, where nodes are represented by yellow dots. This tree provides a representation of the evolutionary relationships between different strains. We use this phylogenetic information to inform our data split, ensuring that closely related strains are not separated between the training and test sets. Specifically, we implement a monophyletic clade-based splitting strategy, where entire clades below a certain depth in the tree are assigned to either the training or test set. This step is crucial to prevent information leakage and maintain the integrity of our evaluation.

Furthermore, we implement a sequence similarity cutoff of 95\% using the Levenshtein distance metric to address the issue of high sequence homology between training and test sets. Sequences with greater than 95\% similarity are grouped together and assigned entirely to either the training or test set, never split between the two. In total, we create 5000 parent-child sequence pairs for training and 500 pairs for testing. 

\section{ Variances for the Nucleotide Transformer Benchmarks}

Table \ref{table:ntbenchmarksvariances} shows the standard deviations of the 10-fold  Matthews Correlation Coefficient (MCC) scores for the Nucleotide Transformer (NT) benchmarks. The peer-reviewed baselines are sourced from a leaderboard maintained by the authors of \cite{DallaTorre2023TheNT} on the Hugging Face platform \cite{leaderboard}.

\begin{table*}
	\centering
	\begin{center} 
	\begin{tabular}{ c | c | c c | c }
	
	\toprule
	
	\textbf{NT Benchmark}  & \begin{tabular}{@{}c@{}}Peer-reviewed \\ Baselines \end{tabular}   &  \begin{tabular}{@{}c@{}}ENBED \\ (GRCh38)\end{tabular}   & ENBED  &  Std. Dev. \\
	\midrule
	H3                    & 0.791 \cite{DallaTorre2023TheNT}       	         & 0.723 &{0.802}  & 0.031  \\
	H3K14ac               & 0.612 \cite{Nguyen2023HyenaDNALG}         	 		& 0.537 & {0.636} & 0.020   \\
	H3K36me3              & 0.616 \cite{DallaTorre2023TheNT}       	         & 0.611 & {0.624}  & 0.016  \\ 
	H3K4me1               & 0.544 \cite{DallaTorre2023TheNT}       	         & 0.498 & {0.591}  & 0.009  \\
	H3K4me2               & 0.455 \cite{Nguyen2023HyenaDNALG}          		& 0.433 & {0.501} & 0.035   \\
	H3K4me3               & 0.549 \cite{Nguyen2023HyenaDNALG}          		& 0.580 & {0.587} & 0.018   \\
	H3K79me3              & 0.672 \cite{Nguyen2023HyenaDNALG}          		& 0.648 & {0.756} & 0.014   \\
	H3K9ac                & 0.581 \cite{Nguyen2023HyenaDNALG}          		& 0.427 & {0.590} & 0.006   \\
	H4                    & 0.807 \cite{DallaTorre2023TheNT}       	         & 0.750 & {0.823}  & 0.011  \\
	H4ac                  & 0.564 \cite{Nguyen2023HyenaDNALG}          		& 0.548 & {0.605} & 0.017   \\
	\midrule
	Promotor (all)        & 0.950  \cite{DallaTorre2023TheNT}                & 0.906 & {0.961} & 0.021   \\
	Promotor (non-TATA)   & 0.952  \cite{DallaTorre2023TheNT}                & 0.892 & {0.959} & 0.019   \\
	Promotor (TATA)       & 0.920 \cite{Avsec2021}                  & 0.883 & {0.944} & 0.017   \\
	\midrule
	Splice acceptor       & 0.973 \cite{DallaTorre2023TheNT}         			& 0.754 & 0.943  & 0.034   \\
	Splice donor          & 0.974 \cite{DallaTorre2023TheNT}        			& 0.835 & 0.911  & 0.029   \\
	\midrule
	Enhancer              & 0.548 \cite{DallaTorre2023TheNT}                 & 0.577 & {0.585} & 0.011   \\
	Enhancer Types        & 0.450 \cite{DallaTorre2023TheNT}                  & 0.459 & {0.482} & 0.007   \\
	\bottomrule
	\vspace{2mm}
	\end{tabular}
	
	\caption{\textbf{Nucleotide Transformer (NT) Variances.}}
	
	\label{table:ntbenchmarksvariances}
	\end{center}
	\end{table*}

\section{Evaluation Metrics}

\subsection{Matthew's Correlation Coefficient}

The Matthews Correlation Coefficient (MCC), originally introduced by Matthews in 1975 for binary classification, has been extended to multi-class classification scenarios \cite{Gorodkin2004}. In the multi-class context, the MCC provides a balanced measure of the quality of classification that is particularly useful when dealing with imbalanced datasets. For a classification problem with K classes, the multi-class MCC is defined as:
\begin{equation}
MCC = \frac{c \times s - \sum_{k} p_k \times t_k}{\sqrt{(s^2 - \sum_{k} p_k^2) \times (s^2 - \sum_{k} t_k^2)}}
\end{equation}
where c is the total number of correctly classified samples, s is the total number of samples, $p_k$ is the number of times class k was predicted, and $t_k$ is the number of times class k truly occurred. The coefficient yields values in the interval [-1, 1], with 1 indicating perfect prediction, 0 signifying random prediction, and -1 denoting complete misclassification.
The MCC takes into account all elements of the confusion matrix, providing a more comprehensive evaluation than metrics such as accuracy or F1-score, especially for imbalanced datasets. The MCC also remains informative even when class sizes differ significantly and is sensitive to both over-prediction and under-prediction of classes.

We use the scikit-learn implementation of the MCC for our evaluation, which is available in the module: \begin{verbatim}sklearn.metrics.matthews_corrcoef\end{verbatim}

\section{ Accuracy-based Evaluation of the Benchmarks}

Table \ref{table:ntbenchmarks} shows the 10-fold mean accuracy (\%) scores of the best performing variants of the Enformer \cite{Avsec2021}, DNABERT \cite{zhou2023dnabert2}, Nucleotide Transformer v2 \cite{DallaTorre2023TheNT}, and HyenaDNA \cite{Nguyen2023HyenaDNALG} on the Nucleotide Transformer (NT) benchmarks. The scores are sourced from a leaderboard maintained by the authors of \cite{DallaTorre2023TheNT} on the Hugging Face platform \cite{leaderboard}.

\begin{table*}[ht]
	\centering
	\setlength{\tabcolsep}{4pt}
	\centering
	\begin{center} 
	\begin{tabular}{c c c c c | c c}
	
		\hline
		\noalign{\vskip 1mm}
		\textbf{Genomic Benchmark}&CNN & DNABERT & GPT  &\begin{tabular}{@{}c@{}}HyenaDNA \\ \cite{Nguyen2023HyenaDNALG}\end{tabular} & \begin{tabular}{@{}c@{}}ENBED \\ (no pre-training)\end{tabular} & ENBED \\
		\noalign{\vskip 1mm}
		\hline
		\noalign{\vskip 1mm}
	
	Mouse Enhancers            & 69.0     & 66.9   & 80.1         & \underline{85.1}  & 75.5 &  \textbf{90.3}             \\
	Human Enhancers (Cohn)     & 69.5     & \underline{74.0}   & 70.5         & \textbf{74.2}  & 54.3 & {71.2}            \\
	Human Enhancers (Ensembl)  & 68.9     & 85.7   & 83.5         & \underline{89.2}  & 83.3 & \textbf{92.2}              \\
	Coding vs Intergenomic     & 87.6     & \underline{92.5}   & 88.8 & 91.3          & 84.2 & \textbf{93.0}              \\
	Human vs Worm              & 93.0     & {96.5}   & 95.6         & \underline{96.6}  & 90.8 & \textbf{97.3}            \\
	Human Regulatory Elements          & \underline{93.3} & 88.1 & 91.5         & \textbf{93.8} & 80.8 &  90.2            \\
	Human Promoter (Non-TATA)   & 84.6    & 85.6     & 87.7         & \underline{96.6}  & 83.4 & \textbf{97.2}            \\
	Human OCR (Ensembl) & 68.0       &    75.1   & 73.0         & \underline{80.9}  & 64.3 & \textbf{81.9}                \\
		\noalign{\vskip 1mm}
		\hline
		\noalign{\vskip 2mm}
	\end{tabular}
	
	\caption{\textbf{Genomic Benchmarks.} Accuracy (\%) scores of the \textbf{best} and \underline{second-best} model in the Genomic Benchmarks datasets \cite{Greov2022GenomicBA}. The baseline CNN and GPT scores was calculated by the authors of \cite{Greov2022GenomicBA} and \cite{Nguyen2023HyenaDNALG} respectively.}
	
	\label{table:genbenchmarks}
	\end{center}
\end{table*}

\begin{table*}
\centering
\begin{center} 
\begin{tabular}{c c c c c | c c}

	\hline
	\noalign{\vskip 1mm}

\textbf{NT Benchmark} & Enformer  & DNABERT-2 & NT (v2)   & HyenaDNA  & \begin{tabular}{@{}c@{}}ENBED \\ (no pre-training)\end{tabular}  & ENBED  \\
\noalign{\vskip 1mm}
	\noalign{\vskip 1mm}
    \hline
	\noalign{\vskip 1mm}
H3                    & 85.9     & 89.3      & \underline{89.5}                     & 88.9         & 64.4 &\textbf{90.6} \\
H3K14ac               & 63.5     & 75.9      & 76.9                             & \underline{80.9} & 51.6 &\textbf{81.4} \\
H3K36me3              & 67.1     & 79.7      & \underline{81.3}                     & 80.8         & 61.1 &\textbf{82.7} \\
H3K4me1               & 64.6     & 75.8      & \underline{77.7}                     & 75.8         & 58.4 &\textbf{77.9} \\
H3K4me2               & 63.0     & 68.0      & 67.6                             & \underline{73.9} & 55.9 &\textbf{75.7} \\
H3K4me3               & 56.5     & 67.3      & 69.5                             & \underline{77.5} & 50.9 &\textbf{77.9} \\
H3K79me3              & 74.7     & 80.7      & 81.3                             & \underline{83.7} & 83.1 &\textbf{85.4} \\
H3K9ac                & 70.8     & 77.1      & 78.0                             & \underline{79.3} & 60.2 &\textbf{82.6} \\
H4                    & 86.6     & 89.9      & \underline{90.5}                     & 88.2         & 74.3 &\textbf{91.8} \\
H4ac                  & 63.8     & 73.1      & 74.9                             & \underline{78.4} & 67.2 &\textbf{80.5} \\
	\noalign{\vskip 1mm}
	\hline
	\noalign{\vskip 1mm}
Promotor (all)        & 95.4     & 97.1      & \underline{97.6}                     & 96.0         & 94.3 &\textbf{98.0} \\
Promotor (non-TATA)   & 95.5     & 97.2      & \underline{97.6}                     & 96.0         & 94.4 &\textbf{98.0} \\
Promotor (TATA)       & 96.0     & 95.5      & \underline{96.6}                     & 94.1         & 92.9 &\textbf{96.8} \\
	\noalign{\vskip 1mm}
	\hline
	\noalign{\vskip 1mm}
Splice acceptor       & 91.4     & \underline{97.5}      & \textbf{98.7}                    & {95.8} & 87.8 &{95.8}  \\
Splice donor          & 90.6     & \underline{96.3}      & \textbf{98.7}                    & {95.8} & 87.7 &95.4          \\
	\noalign{\vskip 1mm}
	\hline
	\noalign{\vskip 1mm}
Enhancer              & 72.3     & 75.7      & \underline{77.3}                     & 75.9         & 65.2 &\textbf{78.3} \\
Enhancer Types        & 55.4     & 62.0      & \underline{62.6}                     & 59.5         & 51.4 &\textbf{70.0} \\

	\noalign{\vskip 1mm}
	\hline
	\noalign{\vskip 2mm}

\end{tabular}

\caption{\textbf{Nucleotide Transformer (NT) Benchmarks.} }

\label{table:ntbenchmarks}
\end{center}
\end{table*}

\end{appendices}

\end{document}